# On Image Filtering, Noise and Morphological Size Intensity Diagrams

Vitorino Ramos, Fernando Muge

CVRM - Centro de Geosistemas, Instituto Superior Técnico
Av. Rovisco Pais, 1096, Lisboa Codex,
PORTUGAL
{vitorino.ramos, muge@alfa.ist.utl.pt}

**Abstract:** In the absence of a pure noise-free image it is hard to define what noise is, in any original noisy image, and as a consequence also where it is, and in what amount. In fact, the definition of noise depends largely on our own aim in the whole image analysis process, and (perhaps more important) in our self-perception of noise. For instance, when we perceive noise as disconnected and small it is normal to use MM-AF filters to treat it. There is two evidences of this. First, in many instances there is no *ideal* and pure noise-free image to compare our filtering process (nothing but our self-perception of its pure image); second, and related with this first point, MM transformations that we chose are only based on our self - and perhaps - fuzzy notion. This also yields a third point: that once the *appropriate* filter is found, it is no longer applicable for a new noisy image, with a different kind of noise intensity, distribution and size. In other words, the design of MM filtering algorithms for one particular noise-removal problem and by using our perception is only extended for similar images. Algorithm robustness and adaptation is no longer possible. However, in the absence of that *ideal* pure noise-free image and by using the strategy of comparing two simultaneous filtering process on the same original noisy image, it is possible to find some relations that can help us, one step more through the direction of automatically chose the *right* filtering process. The present proposal combines the results of two MM filtering transformations ($FT_1$, $FT_2$) and makes use of some measures and quantitative relations on their Size/Intensity Diagrams to find the *most* appropriate noise removal process. Results can also be used for finding the most appropriate stop criteria, and the right sequence of MM operators combination on Alternating Sequential Filters (ASF), if these measures are applied, for instance, on a Genetic Algorithm's target function.

## 1. INTRODUCTION

Mathematical morphology (MM) filters are non-linear filters suited to different filtering tasks. First, a morphological filter may be used for restoring images corrupted by some type of noise. This is an important issue because most image interpretation techniques and measurements are strongly obstructed by the presence of noisy data. A noisy image must therefore be filtered prior to any further processing such as edge detection, segmentation, and grey scale measurements. There exist also many linear filters for linear filtering noisy images but, contrary to non-linear such as morphological filters, they usually fail to preserve edges. Second (and perhaps more important, in the scope of the present work), a morphological filter may be used to selectively remove image structures or objects while preserving the other ones. The selection is based on the geometry and local contrast of the image objects. In this sense, a morphological filter can already be interpreted as a step towards the interpretation of the image (*Soille*, [15]).

This last point arises one important question. Which filtering task is more appropriate for a given image? – or by other words, if MM filters can be seen as a step through image interpretation, which filter is more related to our *common* self-perception of that image, or our self-perception of his noisy parts? As we know, in the absence of a pure noise-free image (for comparison purposes) it is hard to define what noise is, in any original noisy image, and as a consequence also where he is, and in what amount. In fact, the definition of noise depends largely on our own aim in the whole image analysis process, and (perhaps more important) in our self-perception of noise. For instance, when we perceive noise as disconnected and small it is normal to use MM-AF filters to treat it, but are we not *a priori* incorporating our knowledge into the algorithm? There are two evidences of this. First, in many instances there is no *ideal* and pure noise-free image to compare our filtering process (nothing but our self-perception of its pure image); second, and related with this first point, MM transformations that we chose are only based on our self - and perhaps - fuzzy notion. This also yields a third point: that once the *appropriate* filter is find, it is no longer applicable for a new noisy image, with a different kind of noise intensity, distribution, shape and size. In other words, the design of MM filtering algorithms for one particular noise-removal problem and by using our perception is only extended for similar images. Algorithm robustness and adaptation is no longer possible.

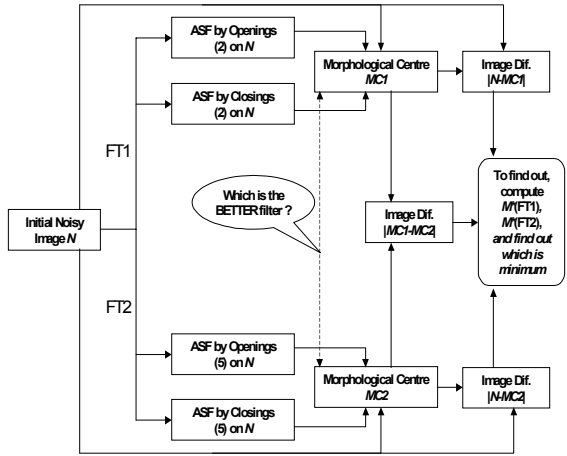

Fig.1 - One original noisy image $N$ is submitted simultaneously to two filtering transforms ($FT_1$ and $FT_2$), and noise measures $M^*$ are computed by comparing results between them (the concept of *relative* cleaning efficiency is incorporated). In the present approach, and only has an example, the two filtering processes used the definition of morphological centre (which use Alternating Sequential Filters), with different sizes (2,5; image $MC_1$ is the output of $FT_1$ and image $MC_2$ of $FT_2$).

Robust noise measures are indeed important in many aspects. First, as our common sense suggests, they can be used to find differences on similar images, for instance. Or they can be used to identify regions where noise can occur with more or less intensity, or even to measure the successful application in many filtering transformations. But perhaps more interesting is to use this potential to - by using mathematical morphology operators - automatically chose the most appropriate filtering transformation for a given image. In fact, the MM filter can adapt itself to that image, or evolve better solutions by using any optimisation algorithm (e.g. Conjugate-gradient methods, Simulated Annealing, Genetic Algorithms), using these measures as his objective functions. For example, if we strictly speak about alternating sequential filters, first proposed by Sternberg (ASF – [16]; *Maragos* and *Schafer*, [7]; *Serra*, [12,13,14]; *Matheron*, [8]; *Meyer* and *Serra*, [10,11]) there are at least two algorithm instances that must be found - when should we stop the process, and what morphological operators sequential combination is the most appropriate (i.e. the algorithm structure). This kind of approach was in fact followed by *Huttunen et al* (1992, [4]) or more recently by *Harvey*, *Marshall* and *Kraft*, using Genetic Algorithms ([1], [2,3], Kraft *et al*, [5]). On applying the GA to some noise-reduction problems, *Harvey et al* used a 100 x 100 portion of the original uncorrupted image (*Lena*) and a corresponding 100 x 100 pixel portion of a corrupted image as a set test. The best soft morphological filter found up to that point (1000 generations) was then applied to the full-size corrupted image. Several error criterions (mean absolute errors, mean squared errors, signal-to-noise ratios, and peak signal-to-noise ratios) of the corrupted and the filtered images, with respect to the original uncorrupted images, were then computed at each generation, to quantify the filters performance found by the GA.

Another kind of MM adaptation was also tried by *Salembier* (1992), developing structuring element (SE) adaptation for filtering noisy images. In his approach, the SE is tuned at each image pixel so as to minimise the difference between the input and desired signals. Once again, the overall strategy involves combinatorial optimisation. These ideas can be used for instance, on noise removal in ancient book page cleaning, if some uncorrupted pages (processed by human hand) are applied for discovering the appropriate filtering process to the rest of the book (to other types of approaches see for instance, *Mengucci et al*, [9]). There is however, a drawback in *Salembier* and *Harvey* models: there must exist a target image (or a portion of it), condition that in most real world cases can not be fulfilled.

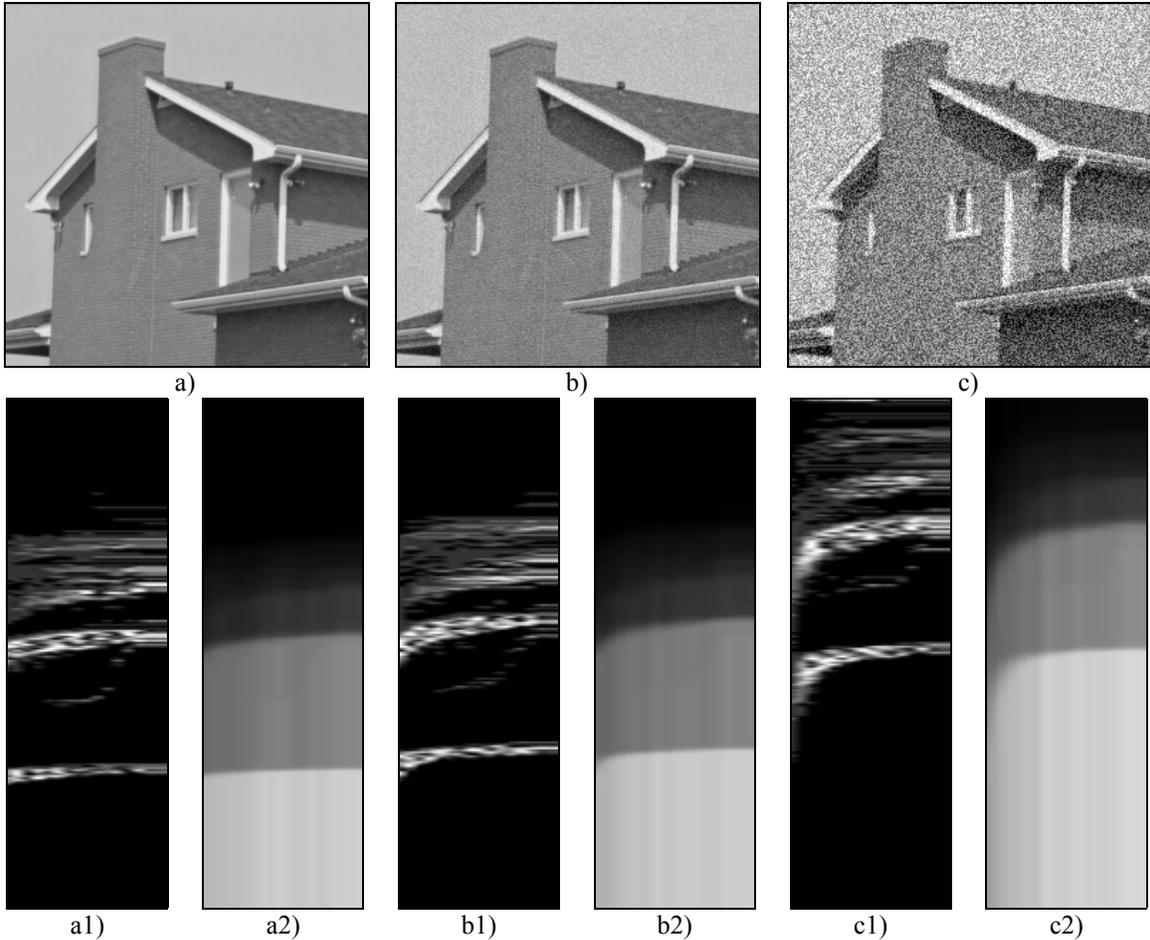

Fig.2 – **a)** Original *House* 8 bit image (256x256 pixels). **b-c)** The same image added with 10%, 50% random *Salt and Pepper* noise, respectively. **a1-b1-c1)** Global histograms of the original images (a-b-c) and from 15 openings, represented as an image - each column $r$ represents an histogram from the original image opened with size $r$, and on each row $k$ the value of different histograms at grey level $k$ (images were multiplied by 5 in width and by 2 in height; pixel values – or frequency - were multiplied by 50 - 0,0 co-ordinates are at the upper left corner). **a2-b2-c2)** Size/Intensity Diagrams of the original images (a-b-c). Again each column $r$ represents the cumulated histogram for the original image opened with size $r$, and each row $k$ the value of different cumulated histograms at grey level $k$ (images were multiplied by 5 in width and by 2 in height).

## 2. INCORPORATING RELATIVITY

One new and possible direction is to treat the concept of a free-noisy image as an idealisation that provides a useful reference point. For example, one can ask (instead) about the *relative* cleaning efficiency of noisy images with respect to each other, e.g., one image cleaning process against other, one particular cleaning strategy in one image region against the same process in another region, etc. The advantages of the concept of relative efficiency, as opposed to the all-or-nothing notion of absolute efficiency, are easy to spot by way of an analogy to the concept of efficiency as used in physics. Heat engines can be given an efficiency rating based on the fraction of available energy that they convert into useful work. A refrigerator with an efficiency of 40% might be considered quite good, and a buyer would prefer this to one with an efficiency of 35%.

Of course, no one would ever expect 100% efficiency. These ideas can be incorporated into our problem if, for instance, we consider two (or more) filtering processes running simultaneously on the same original corrupted image ($FT_1$ and $FT_2$; Figure 1). The present proposed measure (see section 3) is then computed for the two filtering processes ($M^*_1$ and $M^*_2$), by comparing image differences between the outputs of $FT_1$ and $FT_2$, and also with the corrupted input image $N$. As an example, we used the definition of morphological centre (by computing the respective Alternating Sequential Filters), with two different sizes has the two filtering processes. The strategy was then applied in two original corrupted images (with artificially created noise - fig.2 b,c) using image *House* (from USC-SIPI Database - Univ. of Southern California - Signal and Image Processing Institute[http://sipi.usc.edu/services/database/Database.html]), added with 10% and 50% random *salt and pepper* noise.

## 3. QUANTIFYING NOISE

It is clear that under the present theoretical framework, image differences can incorporate important information about the relative efficiency of the two filtering processes. For instance, if we consider the difference between the $FT_1$ output image ($MC_1$) and the original corrupted $N$ image (that is, the $|MC_1 - N|$ image), brighter pixels are indeed more important to take in account, since they reveal bigger differences between the two images. And bigger differences (in this particular case) reveal that the $FT_1$ transform is changing grey level intensities in the original image, in greater amounts (for better or for worse - let us address this problem later). For this reason, the frequency of these type of pixels should be quantified, and naturally this value should be weighted by their respective intensities (since they reveal bigger or lower differences in the respective filtering process). Let us assume for the moment, that we have one noisy $N$ image produced by adding random *salt and pepper* noise on one pure noise-free $F$ image, that we have access to. If we compare both, the result of the quantification of all weighted frequencies should be higher if our noise implantation process was drastic. By drastic, we mean that a great number of pixels were modified in their intensities. In other words, the proposed measure $M$ should be in the form of (Eq.1):

$$M_F = \int_0^k k \cdot H(k)_{|F-N|} dk$$

,where $H(k)$ is the histogram function of the difference image, for all the $k$ grey level intensities. Notice also that the present formulation must be computed in discrete terms. Since $M$ is a global measure between grey level intensity differences among two images, there is however a drawback. One noisy image $N_1$, obtained from his one pure noise-free original, that has two different pixels (2 that changed from 30 to 40 in grey level intensity, in some 2 particular pixels) is as much noisy - by this criteria - as one noisy image $N_2$, that has only one different pixel (1 that changed from 115 to 135, for instance). If we assume any two image filtering transformations ($FT_1$ and $FT_2$) it is possible to compute $M(FT_1)$ and $M(FT_2)$. By taking the smallest value between the two measures it is then possible to get an idea of which transformation is the *better* (assuming, that *better* represents any good noise removal image transformation, and taking also in account that noise itself is hard to define, and to localise). Due to the last reason, there is however, two hypothetical problems with the present approach.

The first problem is due to the fact that $M$ is a global measure. Consider for example two filtering transformations $FT_1$ and $FT_2$ operating in one corrupted image $N$. Let us suppose also that $FT_1$ is mainly changing pixels (let us assume, properly) in noise-free parts of that image with a small and reduced series of changes, while $FT_2$ is doing so in the right regions (i.e. in noise regions), globally with more changes. If we take the above strategy it is clear that $FT_1$ will be chosen as the appropriate filtering transform ($M(FT_1) < M(FT_2)$) – being in fact $FT_2$ the right one to pick, since mainly noise regions are being transformed. Or consider even another hypothetical case where any single FT produces one image that is equal to the initial noisy image (i.e. no differences, which implies $M(FT)=0$). This problem arises because it is difficult (if not impossible) to define noise itself. And if it is difficult to define, it is even harder to localise it. As we can see from the above statements, what we want is simply to see *M* obtaining values near zero in free-noisy regions and precisely the opposite for corrupted regions.

By other words, $M \to 0$ when we compute *M* in free-noisy image regions, and $M \to max$ when we compute it in noisy regions (ideally for these regions *M* should be in the order of the grey level differences between, only pure and noisy pixels). The problem is that when we compute *M* in the presence of a pure clean initial image, all the image comparison (image subtraction) is made within 100% good noise-free parts. On the other hand, when we compute *M* when we have a noisy initial image, some parts of *M* are computed (compared) within good free noisy regions and others do not.

Since it is difficult (if not impossible) to find what and where noise is (that is our aim, after all), another solution must be find. One simple solution could be the following: let us go back to the previous hypothetical situation. If $FT_1$ and $FT_2$ are mainly operating in different regions, it is clear that the $|MC_1 - MC_2|$ image should be very different from an empty set, and if in fact $FT_2$ is the appropriate transformation to take (since it operates mainly within the noisy pixels – (Eq.1) should be transformed. By this argument, instead of picking the FT which have the lowest *M*, we should instead pick the FT with the lowest *M\** (Eq.2):

$$M^*_{|FT|_r} = M_{|N-FT|_r} \cdot M_{|FT_1-FT_2|_r} =$$
$$= \int_0^k k \cdot H(k)_{|N-FT|_r} \cdot dk \cdot \int_0^k k \cdot H(k)_{|FT_1-FT_2|_r} dk$$

The second problem is related to the fact that relationships about shape, neighbourhood and size in noisy regions are not conveniently taken into consideration, only by measuring *M* expressed in equation Eq.1. One possible extension is to compute the proposed measure *M\**, also for different morphological opening images obtained from the first (and/or via morphological closings), that is, for different values of *r* (size of structural element used for any closing and/or opening). A similar approach was first considered by *Lotufo* and *Trettel* (1996, [6]) when they proposed the Size-Intensity (SI) morphological diagram. The idea was that the classical intensity histogram is a grey-scale granulometric curve when using sequences of opening by cylinders of single pixel structuring elements and as the measurement, the projected area of the function. Based on this fact, the two authors proposed a 2D grey-scale granulometry diagram with axis *r* and *k*, where *r* and *k* are the radius and the amplitude (intensity) of the related structuring element, respectively, used in the openings. Each coordinate (*r,k*) of the diagram contains the area of the grey-scale image at each opening. The present proposed measure *M\** is in fact very near with the SI diagram definition, since SI of one image *I(x,y)*, can be defined as (Eq.3):

$$SI(r,k) = \int_0^k H(k)_{I_r} dk$$

It becomes clear, by comparing Eq.1 with Eq.3, that suffices to compute a non-cumulative Size-Intensity diagram for each (*r,k*), i.e. one new image (see figure 2 – a1,b1,c1 / for the classical SI, in *Lotufo* and *Trettel* terms, see instead figure 2 – a2,b2,c2), and to multiply this last image by the correspondent *k* value to extract the value of *M* (if of course, at the end, we compute the sum of all grey-level intensities – i.e. the volume). Similarly, $M^*_{|FT|}$ can be computed if image grey-level intensities of the $k.SI(r,k)_{|N-FT|}$ image, are multiplied by the grey-level intensities of image $k.SI(r,k)_{|FT1-FT2|}$ (notice that these SIs are non-cumulative, and again the volume must be calculated, since *k* is not constant and can not go outside the integral in Eq.1). Classic cumulative Size-Intensity diagrams, in graphic form, can be seen in figure 3 – a2,c2. In figure 4, one can get an idea how $FT_1$ and $FT_2$ are changing the grey-level histogram of the initial corrupted images.

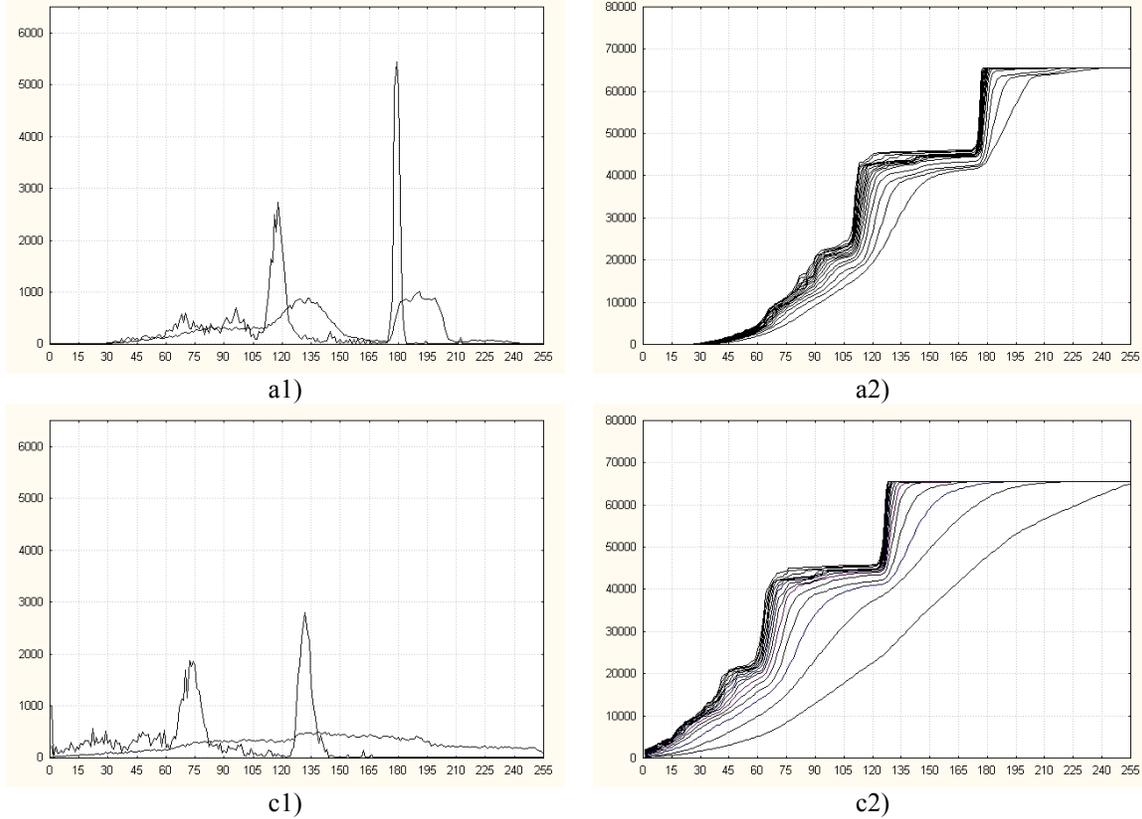

Fig. 3 - **a1-c1)** Histograms of original images (a-c on Fig.2) and for these images (a-c) opened 5 times (upper curves). **a2-c2)** Classical Size/Intensity diagrams for images a and c (Fig.2) respectively, on graphic form (lowest curve is for $r=0$ / upper curve for $r=15$).

## 4. RESULTS AND CONCLUSIONS

Applying the strategy defined in figure 1, we come up with figure 5. For the corrupted 10% image $N$ (fig.2b), the $|N - MC_1|$ and the $|N - MC_2|$ images (fig.5b,5c) are produced applying $FT_1$ and $FT_2$ respectively (fig.1). Then, by comparing $FT_1$ and $FT_2$, the image $|MC_1 - MC_2|$ is computed (fig.5a). From this point, non-cumulative Size-Intensity diagrams are produced (fig.5a1-b1-c1, respectively from fig.5-a-b-c). Only for comparison reasons, the same was done for a different corrupted image $N$ with 50 % noise (fig. 2c; results are in fig. 5d-e-f-d1-e1-f1). At this stage, $M^*_{|FT_1|}$ can be easily computed by multiplying the a1-b1 images by the respective $k$ intensities, and finally by multiplying the volume of these last images. Similarly for $M^*_{|FT_2|}$ the method is the same, but with images a1-c1. As we can perceive, results of $M^*_{|FT_1|}$ are lower than $M^*_{|FT_2|}$, since image b1 is darker than c1. Indeed and as pointed by our method, the right FT to chose is $FT_1$ (compare also the images fig.4b-c), as corrupted images were added with a small type of noise, and ASFs with smaller structural element sizes ($FT_1$) are indeed more appropriate.

A bigger effort, however, must be put in analysing the particular contributions of $M_{|N-FT|}$ and $M_{|FT_1-FT_2|}$ (equation 2) on the above strategy. In what conditions, for instance, we should come up with high $M_{|N-FT|}$ values, and if this is relevant, should we consider also the hypothetical cases where $FT_1$ and $FT_2$ are mainly transforming different regions. To our perception, is it really better to come up with a filtering transformation that changes lesser pixels mainly in noisy regions, or to consider another FT that produces the same effect but in higher scale, corrupting slightly and simultaneously noise-free regions? Although, the present approach is manifestly far from the perceptive idea of noise, we believe that it can point out, one step more through a possible way.

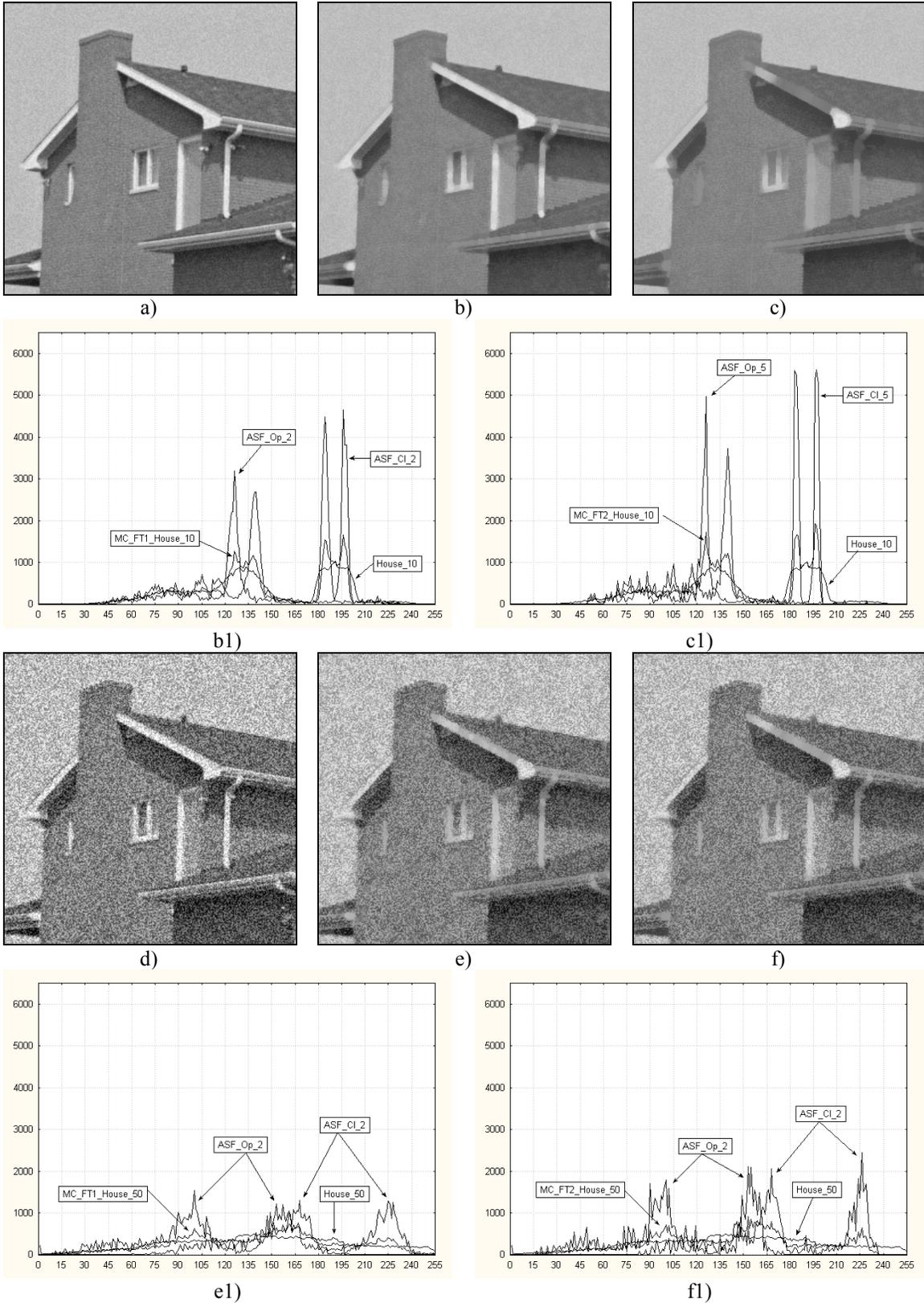

Fig. 4 – **a)** Original corrupted image (*House* with 10% noise). **b)** Morphological Centre from a), computed by 2 ASF starting by openings and closings of size 2 (FT$_1$). **c)** The same as b) but with size 5 (FT$_2$). **b1-c1)** Histograms of FT$_1$ and FT$_2$. **d-e-f-e1-f1)** The same as a-b-c-b1-c1) but for *House* with 50% noise (d).

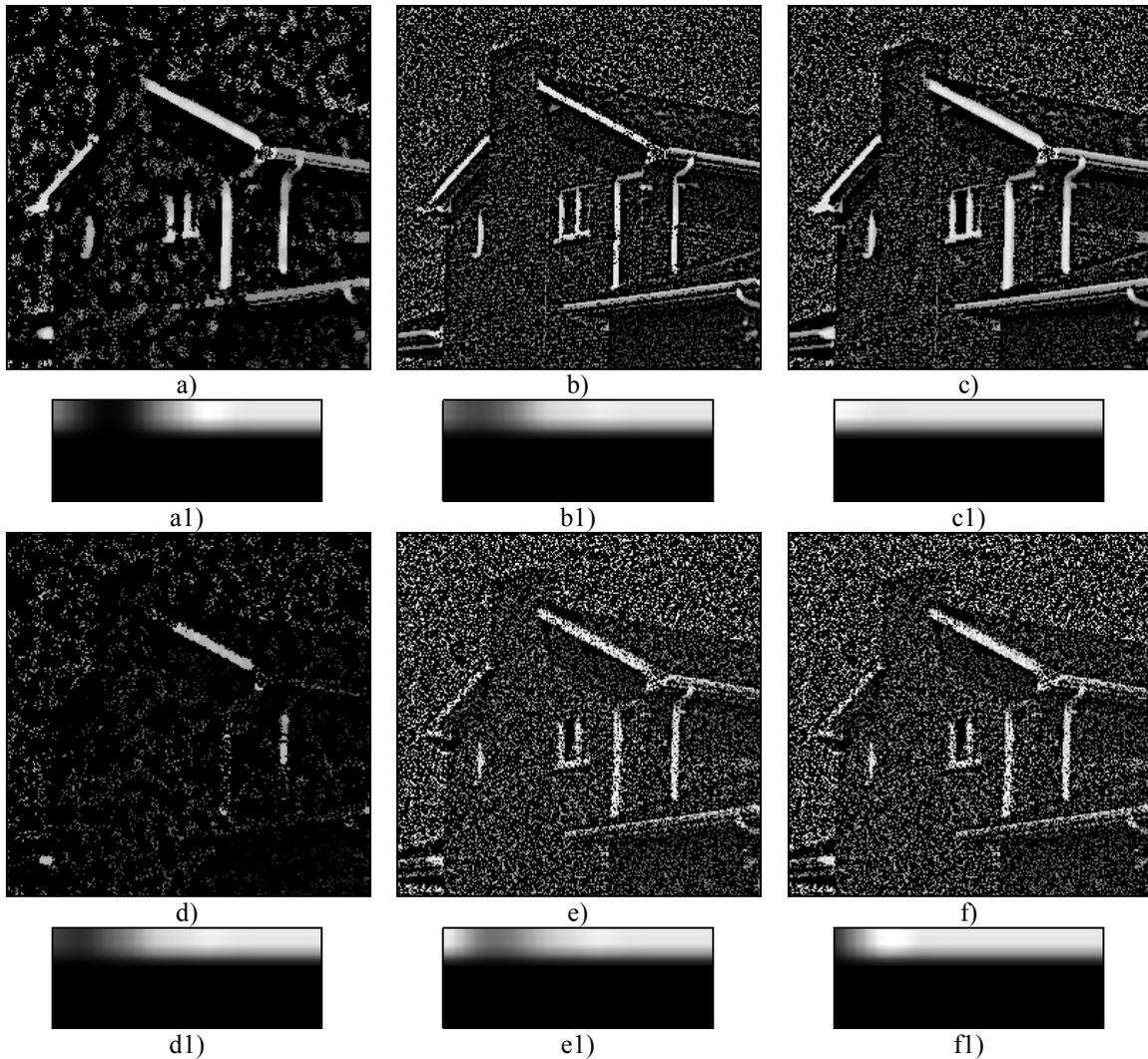

Fig.5 – **a)** MC1-MC2 (from $FT_1$ and $FT_2$) for *House* with 10% noise. **b)** N-MC1 (from $FT_1$) for *House* with 10% noise. **c)** N-MC2 (from $FT_2$) for *House* with 10% noise. **a1-b1-c1)** The respective non-cumulated Size/Intensity Diagrams for a), b) and c). **d-e-f-d1-e1-f1)** The same as a-b-c-a1-b1-c1) but for *House* with 50% noise (images of SIs are zoomed – only the first 5 *r* (openings - columns) and the first 50 *k* (intensities - rows) are showed; images were multiplied by 5 in width and by 2 in height; pixel values – or frequency - in SIs, were multiplied by 50 - 0,0 coordinates are at the upper left corner).

## ACKNOWLEDGEMENTS

The first author wishes to thank to *AMCR*, and to FCT-PRAXIS XXI (BD20001-99), for his PhD Research Fellowship.

## REFERENCES

[1] Harvey, N & Marshall, S. (1994), Using Genetic Algorithms in the Design of Morphological Filters", *in* Serra, J. & Soille, P. (Eds.), "Mathematical Morphology and its Applications to Image and Signal Processing", pp. 53-59, Kluwer Academic Publishers, Boston.

[2] Harvey, N. & Marshall, S. (1996), "The use of Genetic Algorithms in Morphological Filter Design", Signal Processing: Image Communication 8(1), pp. 55-72.

[3] Harvey, N & Marshall, S. (1996), Grey-scale Soft Morphological Filter Optimization by Genetic Algorithms, *in* Maragos, P., Schafer, R. & Butt, M. (Eds.), "Mathematical Morphology and its Applications to Image and Signal Processing", pp. 179-186, Kluwer Academic Publishers, Boston.

[4] Huttunen, H. Kuosmanen, P., Koskinen, L. & Astola, J. (1992), "Optimization of Soft Morphological Filters by Genetic Algorithms", Proc. of Image Algebra and Morphological Image Processing V, San Diego, USA, July 1994, pp. 13-24.

[5] Kraft, P., Harvey, N. & Marshall, S. (1997), "Parallel Genetic Algorithms in the Optimization of Morphological Filters: A General Design Tool", *Journal of Electronic Imaging* 6(4), 504-516.

[6] Lotufo, R.A. & Trettel E. (1996); Integrating Size Information into Intensity Histogram, *in* Maragos, P., Schafer, R. & Butt, M. (Eds.), "Mathematical Morphology and its Applications to Image and Signal Processing", pp. 281-288, Kluwer Academic Publishers, Boston.

[7] Maragos, P. & Schafer, R. (1987), "Morphological Filters*", IEEE Transactions on Acoustics, Speech and Signal Processing* 35(8), pp. 1153-1183.

[8] Matheron, G. (1988), Filters and Lattices, *in* J.Serra., "Image Analysis and Mathematical Morphology. Volume 2: theoretical advances", Academic Press, chapter 6, pp. 115-140.

[9] Mengucci, M., Granado, I., Pinto, R.C. & Muge, F. (2000), "Figure/Text Segmentation and Classification Applied to Ancient Portuguese Books", *submitted to* ICPR'2000 - 15th International Conference on Pattern Recognition, Barcelona, Spain, 3-8 Sep. 2000.

[10] Meyer, F. & Serra, J. (1989), "Contrasts and Activity Lattice", *Signal Processing* 16, 303-317.

[11] Meyer, F. & Serra, J. (1989), Filters: from Theory to Practice, in "Acta Stereologica", Vol. 8/2, Freiburg im Breisgau, pp. 503-508.

[12] Serra, J. (1988), Introduction to Morphological Filters, *in* J.Serra., "Image Analysis and Mathematical Morphology. Volume 2: theoretical advances", Academic Press, chapter 5, pp. 101-114.

[13] Serra, J. (1988), The Centre and Self-dual Filtering, *in* J.Serra., "Image Analysis and Mathematical Morphology. Volume 2: theoretical advances", Academic Press, chapter 8, pp. 159-180.

[14] Serra, J. (1988), Alternating Sequential Filters, *in* J.Serra., "Image Analysis and Mathematical Morphology. Volume 2: theoretical advances", Academic Press, chapter 10, pp. 203-214.

[15] Soille, P. (1999), *Morphological Image Analysis – Principles and Applications*, Springer-Verlag, Berlin Heidelberg.

[16] Sternberg, S. (1986), "Greyscale Morphology", *Computer Graphics and Image Processing*, 35, 333-355.